\documentclass[10pt,twocolumn,letterpaper]{article}

\usepackage[pagenumbers]{cvpr} 

\usepackage{graphicx}
\usepackage{amsmath}
\usepackage{amssymb}
\usepackage{amsthm}
\usepackage{mathrsfs}
\usepackage{booktabs}
\usepackage{bm}
\usepackage{algorithm}
\usepackage{algorithmic}
\usepackage{multirow}
\usepackage{bbm}
\usepackage{tablefootnote}
\usepackage[dvipsnames]{xcolor}

\usepackage[pagebackref,breaklinks,colorlinks]{hyperref}

\usepackage[capitalize]{cleveref}
\crefname{section}{Sec.}{Secs.}
\Crefname{section}{Section}{Sections}
\Crefname{table}{Table}{Tables}
\crefname{table}{Tab.}{Tabs.}

\def\cvprPaperID{1584} 
\def\confName{CVPR}
\def\confYear{2022}

\begin{document}

\title{Self-supervised Image-specific Prototype Exploration \\for Weakly Supervised Semantic Segmentation}

\author{Qi Chen\textsuperscript{1}, Lingxiao Yang\textsuperscript{1}, Jianhuang Lai\textsuperscript{1,2,3} and Xiaohua Xie\textsuperscript{1,2,3\thanks{Corresponding Author}}\\
\textsuperscript{1}School of Computer Science and Engineering, Sun Yat-Sen University, China\\
\textsuperscript{2}Guangdong Province Key Laboratory of Information Security Technology, China\\
\textsuperscript{3}Key Laboratory of Machine Intelligence and Advanced Computing, Ministry of Education, China\\
{\tt\small chenq377@mail2.sysu.edu.cn, \{yanglx9, stsljh, xiexiaoh6\}@mail.sysu.edu.cn}

}
\maketitle

\begin{abstract}
Weakly Supervised Semantic Segmentation (WSSS) based on image-level labels has attracted much attention due to low annotation costs.
Existing methods often rely on Class Activation Mapping (CAM) that measures the correlation between image pixels and classifier weight.
However, the classifier focuses only on the discriminative regions while ignoring other useful information in each image, resulting in incomplete localization maps.
To address this issue, we propose a Self-supervised Image-specific Prototype Exploration (SIPE) that consists of an Image-specific Prototype Exploration (IPE) and a General-Specific Consistency (GSC) loss.
Specifically, IPE tailors prototypes for every image to capture complete regions, formed our Image-Specific CAM (IS-CAM), which is realized by two sequential steps.
In addition, GSC is proposed to construct the consistency of general CAM and our specific IS-CAM, which further optimizes the feature representation and empowers a self-correction ability of prototype exploration.
Extensive experiments are conducted on PASCAL VOC 2012 and MS COCO 2014 segmentation benchmark and results show our SIPE achieves new state-of-the-art performance using only image-level labels.
The code is available at \url{https://github.com/chenqi1126/SIPE}.
\end{abstract}

\section{Introduction}
\label{sec:intro}

Semantic segmentation aims to assign a semantic category label to each pixel in an image, which has been widely applied in autonomous driving \cite{feng2020deep}, medical imaging \cite{tajbakhsh2020embracing} and remote sensing image interpretation \cite{hossain2019segmentation}.
Benefiting from Convolutional Neural Networks (CNNs), semantic segmentation has achieved remarkable progress in fully supervised manner.
However, training a fully supervised segmentation model requires a large number of pixel-level annotations, which is notoriously expensive and time-consuming to collect.
An alternative approach is to learn from weak labels, \eg, image-level labels \cite{ahn2018learning}, bounding boxes \cite{lee2021bbam,zhang2021affinity}, scribbles \cite{lin2016scribblesup,Pan2021Scribble,Xu2021Scribble} and points \cite{bearman2016s,Chen2021Seminar}.
Among these works, image-level labels based Weakly Supervised Semantic Segmentation (WSSS) has enjoyed great popularity within the community.

\begin{figure}[t]
    \centering
    \includegraphics[width=0.47\textwidth]{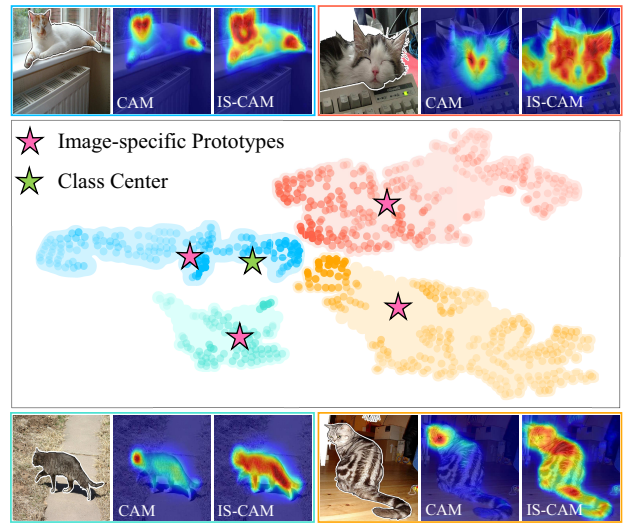}
    \caption{Main motivation. We visualize the pixel-level feature distribution of four cat images by t-SNE \cite{van2008visualizing}. The transparency indicates the magnitude of activation. The original CAM activates each pixel using class center (\textcolor{LimeGreen}{green} star). Our method extracts image-specific prototypes (\textcolor{magenta}{pink} star) to generate Image-Specific CAM (IS-CAM) that captures more complete regions.}
    \label{fig:fig1}
\end{figure}

Most of existing methods leverage Class Activation Mapping (CAM) \cite{Zhou_2016_CVPR} technology to provide localization cues of target object.
Specifically, these methods train a classifier and regard its learned weights as a general representation of each class, \ie, \textbf{class center}.
Then, this class center is used to correlate with image pixels to obtain localization maps as shown in \cref{fig:fig1}.
However, CAM tends to focus on a few primary regions (cat's head) while ignoring other useful cues (cat's body).
To explain the problem, we visualize pixel-level features of foreground extracted from a trained classification network.
Those features are shown with four different colors and their transparency degree indicate the activation of CAM.
We find that the class center always gives high activations to the close pixels (correspond to some primary regions) and ignores the distant pixels.
The imbalanced activations lead to the incomplete localization map as demonstrated in \cref{fig:fig1}.
In addition, activating features on each image by the centroid of that features (pink star) can be beneficial to explore more complete regions (see Image-Specific CAM (IS-CAM) shown in \cref{fig:fig1}).
Therefore, this paper aims to tailor image-specific prototypes to adaptively describe the image itself.

\begin{figure*}[!t]
    \centering
    \includegraphics[width=0.99\textwidth, height=0.31\textwidth]{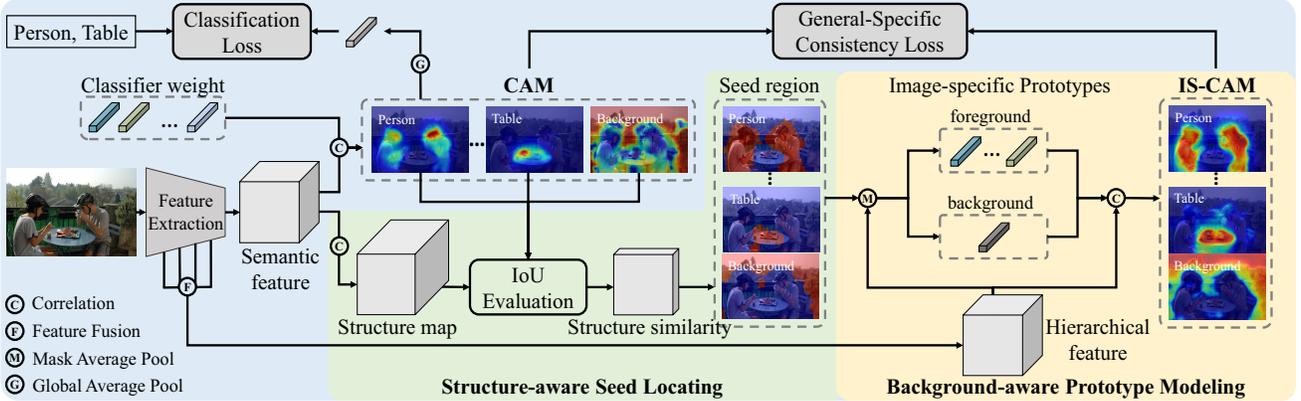}
    \caption{Overview of the proposed SIPE for weakly supervised semantic segmentation. It mainly consists of two proposed methods: an Image-specific Prototype (IPE) Exploration and a General-Specific Consistency (GSC) loss. Specifically, in our IPE, a structure-aware seed locating method is proposed to achieve more robust seed regions and a background-aware prototype modeling is developed to extract hierarchical features. In addition, we add consistent regularization between two types of CAM (\ie, general CAM and our IS-CAM). This self-supervised signal effectively does correction in both CAM and IS-CAM.}
    \label{fig:fig2}
\end{figure*}

To this end, we propose a novel weakly supervised semantic segmentation framework, called \textit{Self-supervised Image-specific Prototype Exploration (SIPE)}.
The proposed SIPE consists of an Image-specific Prototype Exploration (IPE) and a General-Specific Consistency (GSC) loss, which is illustrated in \cref{fig:fig2}.
Specifically, IPE is realized as two sequential steps to characterize prototypes, allowing to capture more complete localization maps.
In the first step, we utilize inter-pixel semantics to explore spatial structure cues, locating robust seed regions of each class.
Given the seed regions, we extract image-specific prototypes and then produce our IS-CAM by prototypical correlation.
In addition, GSC is proposed to construct the consistency of general CAM and our specific IS-CAM.
This self-supervised signal further optimizes the feature representation and empowers a self-correction ability of prototype exploration.
Extensive experiments are conducted on Pascal VOC 2012 \cite{everingham2015pascal} and MS COCO 2014 \cite{lin2014coco} and results show that our SIPE achieves new state-of-the-art performance when only image-level labels are available.

Our main contributions are summarized as:
\begin{itemize}
\vspace{-5pt}
\item
We propose Self-supervised Image-specific Prototype Exploration (SIPE) to learn image-specific knowledge for weakly supervised semantic segmentation.
\vspace{-5pt}
\item
We propose Image Prototype Exploration (IPE) that tailors image-specific prototypes for each image, which is achieved by structure-aware seed locating and background-aware prototype modeling.
It enables the model to capture more complete localization maps.
\vspace{-5pt}
\item
We propose a General-Specific Consistency (GSC) loss to effectively regularize the original CAM and IS-CAM, empowering the feature representation.

\end{itemize}

\section{Related Work}
Due to the low annotation costs, image-level labels based weakly supervised semantic segmentation has attracted increasing attention.
Most existing methods adopt Class Activation Mapping (CAM) to generate localization maps and then refine them as pseudo labels to train a fully supervised segmentation model.
To achieve a high performance segmentation model, many strategies have been investigated to improve the quality of localization maps.

\noindent\textbf{Erasure and accumulation.} Erasure methods explore more object regions by intentionally removing the discriminative regions from the images \cite{wei2017object,Kweon2021unlock,sun2021ecs} or feature maps \cite{hou2018self,choe2020attention}.
However, erasing most of the discriminative regions may confuse the classifier and result in false positives.
To avoid this problem, some works accumulate multiple activations by applying well-designed sampling on dilated convolution rate \cite{wei2018revisiting}, image scales \cite{zhang2020reliability}, spatial location \cite{lee2019ficklenet} and training process \cite{jiang2019integral}.

\noindent\textbf{Cross-image mining.} Considering the sharing semantics between images, some works design cross-image relation mining modules, such as cross-image affinity \cite{fan2020cian}, max bipartite matching \cite{liu2020weakly} and co-attention classifier \cite{sun2020mining} to excavate semantic context of weak labels.
Furthermore, the collaborative information of multi-images is explored to capture the potential knowledge by graph convolution network \cite{li2021group} and self-attention mechanism \cite{wu2021embed}.

\noindent\textbf{Background Modeling.}
Many methods \cite{kim2021discriminative,lee2021railroad,yao2021non,xu2021leveraging} obtain precise background by using auxiliary saliency maps, which is laborious.
Without auxiliary maps, Fan et al. \cite{fan2020learning} propose intra-class discriminator to separate the foreground and the background for each class.
However, due to the object and scene diversity of images, it is quite tricky to learn a general intra-class discriminator for each class.

\noindent\textbf{Self-supervised Learning.} More recent, self-supervised methods mine potential information and build supervisory signals, which has been demonstrated a promising solution to narrow the supervision gap between fully and weakly supervised semantic segmentation.
Wang et al. \cite{wang2020self} apply consistency regularization on CAM from various transformed images to accomplish self-supervision learning.
Chang et al. \cite{chang2020weakly} introduce a self-supervised task that discovers sub-categories, which provides additional supervision to enhance feature representations.

In contrast to existing methods, we fully consider the distinctiveness of images, and introduce image-specific prototypes to discover complete regions and construct a self-supervised manner to empower feature representation.

\section{Approach}
This section elaborates the proposed SIPE framework for weakly supervised semantic segmentation as shown in \cref{fig:fig2}.
Firstly, we briefly review the preliminary of CAM.
Then we describe the pipeline of exploring image-specific prototypes and Image-Specific CAM (IS-CAM).
Finally, a self-supervised learning with General-Specific Consistency (GSC) is introduced to empower feature representation.

\subsection{Class Activation Mapping}
Given an input image and a pretrained classification network, the class activation map $\bm{M_f}\!=\!\{\bm{M}_{k}\}_{k=1}^{K}$ over $K$ foreground classes can be represented as follows:
\begin{equation}
    \bm{M}_k=ReLU({\bm{\theta}_k}^T \bm{F}_{s}), \quad \forall k \in K,
    \label{eq:eq1}
\end{equation}
where $\bm{F}_{s}$ is the semantic feature from the last layer of the network, $\bm{\theta}_k$ denotes the $k$-th classifier weight, and thus $\bm{M}_k$ is the $k$-th class-specific activation map.
Following previous works, CAM is further normalized to $[0 , 1]$ by the maximum value along the spatial axes so that it could be regarded as the probability for each class.

Considering the importance of background in segmentation task, we follow \cite{wang2020self} to estimate the background activation map $\bm{M_b}$ based on $\bm{M_f}$.
Since CAM tends to cover object regions partially, the estimated background often contains high responses in foreground regions, which will bring considerable noise.
To reduce such confusion, we weaken the background probability by introducing an attenuation coefficient $\alpha=0.5$:
\begin{equation}
    \bm{M_b}= \alpha (1-\max_{1\leq k \leq K} \bm{M}_{k}).
    \label{eq2}
\end{equation}
We combine the processed background activation map with the foreground activation map as a whole, \textit{i.e.} $\bm{M}=\bm{M_f}\cup \bm{M_b}$, to help model background knowledge.

\subsection{Image-specific Prototype Exploration}
Image-specific prototype is proposed to represent the feature distribution of each class, allowing to capture more complete regions.
Different from the prototype representation in few-shot segmentation \cite{wang2019panet,liu2020part,zhang2020sg}, there is no ground truth pixel-level mask in WSSS.
To explore image-specific prototypes for characterizing a feature distribution, we design an efficient two-step pipeline.
The first step provides robust class-wise seed regions and the second step aggregates these seeds on a comprehensive feature space to achieve accurate image-specific representation.

\paragraph{Structure-aware Seed Locating.}
A straightforward approach to obtain seeds is empirically selecting thresholds for CAM \cite{huang2018weakly}, but it is difficult to use a fixed threshold on different images due to the diversity of objects and scenarios.
Although CAM pays more attention to discriminative regions, it also produces weak activations on the remaining regions.
It means that CAM has the potential to provide the spatial structure of semantic objects.
Besides, a pixel's spatial structure can be constituted by clustering high correlation pixels.
For an image, we can determine each pixel's category by comparing its spatial structure with CAMs.
Based on above analysis, we propose a structure-aware seed locating method by exploring inter-pixel semantics to capture the spatial structure and employing CAMs as the templates to match the optimal category.

\begin{figure}[t]
    \centering
    \includegraphics[width=0.47\textwidth]{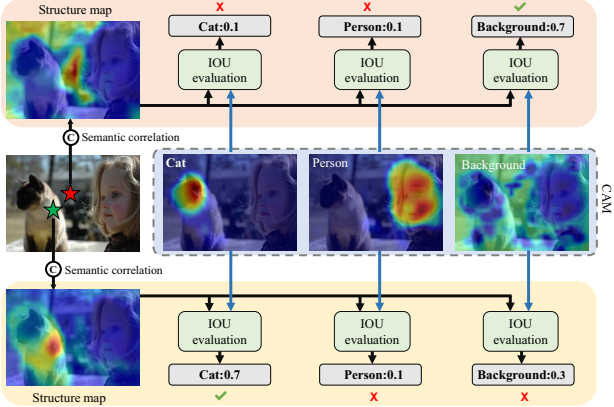}
    \caption{Illustration of structure-aware seed locating of selected foreground (\textcolor{LimeGreen}{green}) and background pixels (\textcolor{red}{red}). The structure map are obtained by semantic correlation. Then the class-wise structure similarity can be evaluated by IoU and the maximal one is selected as the final class.}
    \label{fig:fig3}
\end{figure}

\cref{fig:fig3} illustrates the proposed method with selected foreground and background examples.
Firstly, for an arbitrary pixel $i$, we take its semantic feature vector $\bm{f}^i$ as the query to compute semantic correlation with all pixels in that feature map.
Since pixels with high correlation scores are more likely to belong to the same class, these high correlation pixels can highlight the spatial structure.
Therefore, we define the spatial structure of a pixel by inter-pixel semantic correlation:
\begin{equation}
    \bm{S}^{i}(j) = ReLU(\frac{\bm{f}^i\cdot \bm{F}_s(j)}{||\bm{f}^i|| \cdot ||\bm{F}_s(j)||}),
    \label{eq3}
\end{equation}
where $\cdot$ is dot product and $j$ is spatial index over the feature map as well as the structure map.
$\bm{S}^{i}(j)$ denotes the semantic correlation between pixel $i$ and $j$, and $\bm{S}^{i}$ is the structure map of pixel $i$.
We suppress negative correlations by ReLU function to eliminate the influence of irrelevant pixels.

Secondly, we evaluate the class-wise IoU between the structure map of pixel $i$ and CAM as the structure similarity:
\begin{equation}
    \bm{C}_{k}^{i}=\frac{\sum\nolimits_{j}\bm{M}_{k}{(j)}\bm{S}^{i}{(j)}}{\sum\nolimits_{j}[{\bm{M}_{k}{(j)}}+{\bm{S}^{i}{(j)}} -\bm{M}_{k}{(j)} \bm{S}^{i}{(j)}]}.
    \label{eq4}
\end{equation}
Here, $\bm{C}_{k}^{i}$ denotes the structure similarity for pixel $i$ with respect to $k$-th class.
$j$ is spatial index over the activation map as well as the structure map.
From \cref{fig:fig3}, we can see that the foreground pixel (green star) correlates with the cat's body and achieves the highest IoU with CAM of the cat category.
Additionally, the background pixel (red star) is not associated with foreground pixels, so it is more likely to belong to the background class.

Finally, pixel $i$ is assigned to the category with maximal similarity:
\begin{equation}
    \bm{R}_{k}^{i}=\left\{
    \begin{aligned}
    1,   &\quad\text{if $k=\arg\max_{k'}\bm{C}_{k'}^{i}$},\\
    0,   &\quad\text{otherwise}.
    \end{aligned}
    \right.
\label{eq5}
\end{equation}
By repeating this process for all pixels of the image in parallel, the seed regions $\bm{R}$ of both foreground and background classes are located as shown in \cref{fig:fig2}.

\begin{figure}[t]
    \centering
    \includegraphics[width=0.47\textwidth]{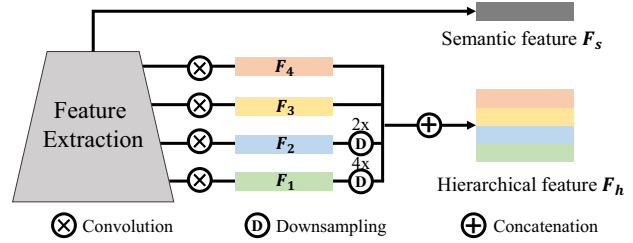}
    \caption{Illustration of the modified backbone for extracting hierarchical feature representations. The semantic feature is extracted from the last layer and the hierarchical feature is the fusion of four stages of the backbone.}
    \label{fig:fig4}
\end{figure}

\paragraph{Background-aware Prototype Modeling.}
In this section, we simultaneously model foreground and background prototypes.
Considering that background does not have specific semantics, it is difficult to explore representative background prototypes on semantic feature space.
Instead, features from shallow layers contain rich low-level visual information (\textit{e.g.} color, texture), which is more suitable to model background-related information.
Therefore, we modify the architecture of the backbone to capture hierarchical features for effective prototype representation.

\cref{fig:fig4} illustrates the architecture of modified backbone.
Specifically, we add four convolution layers to extract multi-scale outputs.
Then the multi-scale outputs are resized to the same size and concatenated to form the hierarchical feature $\bm{F_h}$.
Thus image-specific prototypes $\bm{P}_k$ of foreground and background can be formulated as the centroid of seed regions in hierarchical feature space:
\begin{equation}
    \bm{P}_k = \frac{\sum\nolimits_{i}{\bm{F_h}^{i}*\mathbbm{1}(\bm{R}_k^{i}=1)}}{\sum\nolimits_{i}\mathbbm{1}(\bm{R}_k^{i}=1)},
    \label{eq6}
\end{equation}
where $i$ indexes the spatial locations and $\mathbbm{1}(\cdot)$ outputs 1 if the argument is true or 0 otherwise.
This process performs class-wise compression on the seed pixels, achieving $K$ foreground prototypes and one background prototype.

With these image-specific prototypes, the Image-Specific CAM (IS-CAM) is computed as follows:
\begin{equation}
    \tilde{\bm{M}}_{k}(j) = ReLU(\frac{\bm{F_h}(j) \cdot \bm{P}_k}{|| \bm{F_h}(j) ||\cdot || \bm{P}_k ||}),
    \label{eq7}
\end{equation}
where $\tilde{\bm{M}}_{k}(j)$ is the $k$-th IS-CAM at pixel $j$.
The correlation is bounded in $[-1,1]$, and followed by ReLU to remove negative correlations.

Compared to the original CAM that takes classifier weight as the class center to compute the correlation of each pixel, the proposed IS-CAM utilizes prototypes tailored for every image to achieve more complete object regions.
Besides, background prototype modeling provides high-quality background localization cues, which in turn help determine accurate foreground regions.

\subsection{Self-supervised Learning with GSC}
To further take advantage of image-specific knowledge, we introduce a self-supervised learning paradigm.
The overall training loss consists of multi-label classification loss and the General-Specific Consistency (GSC) loss,
\begin{equation}
    \mathcal{L}_{total} = \mathcal{L}_{cls} + \mathcal{L}_{gsc}.
\end{equation}

The classification loss is computed by a multi-label soft margin loss between the image-level category label ${y}$ and the prediction $\hat{y}$, which is obtained by averaging over foreground maps generated by CAM.
\begin{equation}
    \mathcal{L}_{cls} = \frac{1}{K}\sum_{i=1}^{K}{y_{i}\log\sigma(\hat{y}_{i}) + (1-y_{i})\log(1-\sigma(\hat{y}_{i}))},
    \label{eq8}
\end{equation}
where $\sigma$ is the sigmoid activation function.

The GSC is employed to minimize the difference between original CAM activated by classifier weight and IS-CAM activated by image-specific prototypes.
The mathematical definition of this consistency regularization is formulated as the L1 normalization of the two kinds of CAM:
\begin{equation}
    \mathcal{L}_{gsc} = \frac{1}{K+1}||\bm{M} - \tilde{\bm{M}}||_1,
    \label{eq8}
\end{equation}
where $\bm{M}, \tilde{\bm{M}}$ denotes the original CAM and IS-CAM respectively.
The loss is averaged over $K$ foreground classes and one background class.

With this consistency, the image-specific knowledge is injected into the feature representation, and the collaborative optimization is accomplished in the training cycles.
IS-CAM forces the original CAM to pay attention to the absent object regions, which implicitly narrows the feature distance between the discriminative and missing pixels.
Besides, the enhanced semantic and hierarchical features are favorable to capture more comprehensive and accurate image-specific prototypes and improve the quality of localization maps.

\begin{figure*}[!t]
    \centering
    \includegraphics[width=\textwidth]{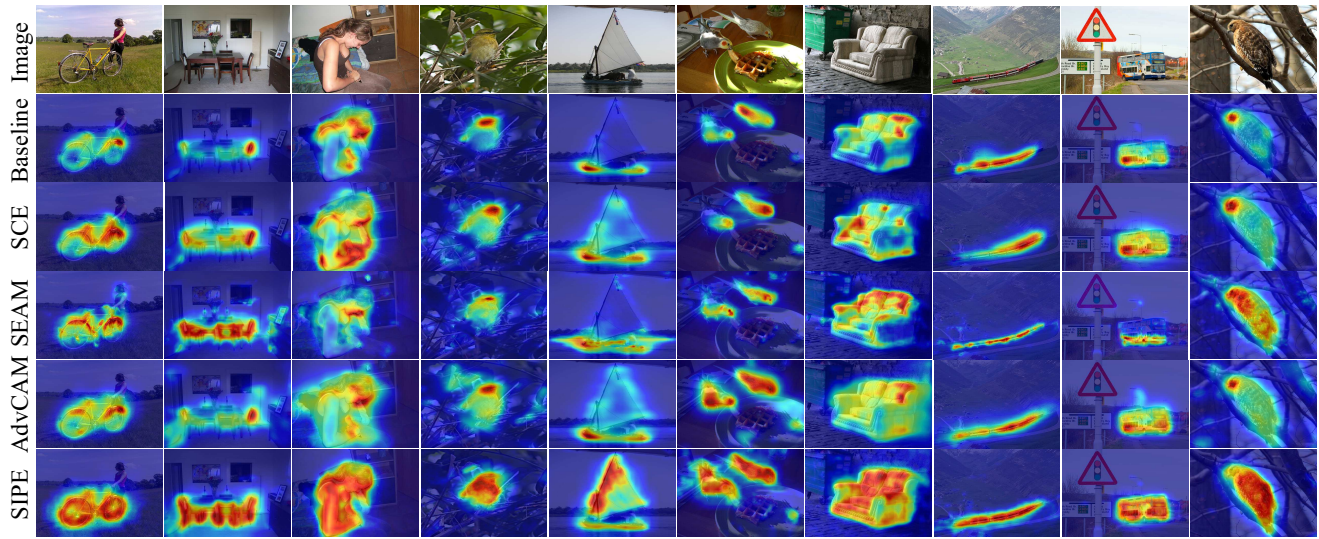}
    \caption{Visual comparison of localization maps generated by different methods on PASCAL VOC 2012 train set. From top to down: original image, Baseline, SCE \cite{chang2020weakly}, SEAM \cite{wang2020self}, AdvCAM \cite{lee2021anti} and our SIPE.}
    \label{fig:fig5}
\end{figure*}

\section{Experiments}
In this section, we first elaborate the experimental settings including dataset, evaluation metric and implementation details.
Second, we compare our method with state-of-the-art approaches on the PASCAL VOC 2012 dataset \cite{everingham2015pascal} and MS COCO 2014 dataset \cite{lin2014coco}.
Third, we conduct a series of ablation studies to verify the effectiveness of the proposed method.

\subsection{Experimental Settings}
\paragraph{Dataset and Evaluated Metric.}
We evaluate our proposed method on PASCAL VOC 2012 segmentation benchmark \cite{everingham2015pascal} with 20 foreground classes and one background class.
The official dataset split contains 1,464 images for training, 1,449 for validation and 1,456 for testing.
Following the common experimental protocol in semantic segmentation, we take additional annotations from SBD \cite{hariharan2011semantic} to build an augmented training set with 10,582 images.
Another MS COCO 2014 dataset has totally 81 classes and contains 80k train and 40k validation images, which is challenging for weakly supervised semantic segmentation.
Note that only image-level classification labels are available during network training for both datasets.
Mean intersection over union (mIoU) is used as a metric to evaluate segmentation results.
The results for the PASCAL VOC test set are obtained from the official evaluation server.

\paragraph{Implementation Details.}
In our experiments, the ImageNet \cite{deng2009imagenet} pretrained ResNet50 \cite{he2016deep} is adopted as backbone with output stride of 16, where fully connected layer is replaced by a classifier with output channels of 20.
The augmentation strategy is the same to \cite{lee2021anti}, including random flipping, random scaling and crop.
The model is trained with a batch size of 16 on 2 Nvidia A100 GPUs.
SGD optimizer is adopted to train our model for 5 epochs, with a momentum of 0.9 and a weight decay of 1e-4.
The learning rates for the backbone and the newly added layers are set as 0.1 and 1, respectively.
We use poly learning scheduler decayed with a power of 0.9 for the learning rate.

\paragraph{Inference}
At the inference stage, the network generates foreground and background seeds by hierarchical features and activates to localization maps.
Instead of checking various mIoU scores over the training set to obtain pseudo labels as other works \cite{wang2020self,lee2021anti}, we directly compute pseudo labels with the background localization maps.

\subsection{Comparison with State-of-the-arts}
\paragraph{Improvements on localization maps.}
We first evaluate mIoU on localization maps, where those maps are generated by the proposed IS-CAM.
\cref{tab:tab1} presents the comparison with other advanced methods on PASCAL VOC 2012 train set.
Among these compared methods, ECS \cite{sun2021ecs} provides the best results with a mIoU of 56.6\%.
Our proposed SIPE achieves the state-of-the-art performance of 58.6\%.
Furthermore, we report the performance with denseCRF post-processing.
The results show that our SIPE with denseCRF improves the mIoU to 64.7\% and outperforms all other methods, which may be benefited from our high-quality localization maps.
We interpret this performance gain comes from the capability of SIPE to generate complete localization maps for foreground and background.
As the localization maps can capture satisfying boundaries of semantic objects, the denseCRF is less confusing while refining the CAMs.
\cref{fig:fig5} shows the visual comparison of foreground localization maps on PASCAL VOC 2012 train set.
It can be observed that our SIPE is effective in capturing the whole semantic regions under various scenes, such as different object scales, crowded objects, and multiple categories.
The obtained high-quality localization maps will further benefit our segmentation results.

\setlength{\tabcolsep}{1.5mm}{
\begin{table}[t]
\caption{mIoU (\%) of localization maps on PASCAL VOC 2012 \textit{train} set. The best results are shown in bold.}
\centering
\begin{tabular}{lccc}
\toprule[1pt]
Method & Pub. & Local. Maps & +denseCRF \\
\midrule[0.5pt]
Baseline    &  - & 50.1 & 54.3  \\
SCE\cite{chang2020weakly}   &  CVPR20 & 50.9 & 55.3  \\
SEAM\cite{wang2020self}  &  CVPR20 & 55.4 & 56.8  \\
EDAM\cite{wu2021embed}  & CVPR21 & 52.8  & 58.2 \\
AdvCAM\cite{lee2021anti} & CVPR21 & 55.6 & 62.1 \\
ECS\cite{sun2021ecs} & ICCV21 & 56.6 & 58.6 \\
CSE\cite{Kweon2021unlock} & ICCV21 & 56.0 & 62.8 \\
SIPE (Ours)  &   & \textbf{58.6} & \textbf{64.7} \\
\bottomrule[1pt]
\end{tabular}
\label{tab:tab1}
\end{table}}

\renewcommand\arraystretch{1.0}
\setlength{\tabcolsep}{0.7mm}{
\begin{table}[!t]
\centering
\caption{Comparison with other state-of-the-arts on PASCAL VOC 2012 \textit{val} and \textit{test} sets. The best results among methods only using image-level labels are shown in bold.}
\begin{tabular}{lccccc}
\toprule[1pt]
Model                & Pub.      & Sup.                  & Backbone   & Val           & Test \\
\midrule[0.5pt]
DSRG\cite{huang2018weakly}       &CVPR18   & $\mathcal{I}$ + $\mathcal{S}$  & ResNet101 & 61.4 & 63.2 \\
SeeNet\cite{hou2018self}         &NIPS18   & $\mathcal{I}$ + $\mathcal{S}$  & ResNet101 & 63.1 & 62.8 \\
FickleNet\cite{lee2019ficklenet} &CVPR19   & $\mathcal{I}$ + $\mathcal{S}$  & ResNet101 & 64.9 & 65.3 \\
OAA+\cite{jiang2019integral}     &ICCV19   & $\mathcal{I}$ + $\mathcal{S}$  & ResNet101 & 65.2 & 66.4 \\
G-WSSS\cite{li2021group}         &AAAI21   & $\mathcal{I}$ + $\mathcal{S}$  & ResNet101 & 68.2 & 68.5 \\
NSROM\cite{yao2021non}           &CVPR21   & $\mathcal{I}$ + $\mathcal{S}$  & ResNet101 & 70.4 & 70.2 \\
EPS\cite{lee2021railroad}        &CVPR21   & $\mathcal{I}$ + $\mathcal{S}$  & ResNet101 & 71.0 & 71.8 \\
AuxSegNet\cite{xu2021leveraging} &ICCV21   & $\mathcal{I}$ + $\mathcal{S}$  & ResNet38  & 69.0 & 68.6 \\
\midrule[0.5pt]
IRN\cite{ahn2019weakly}          &CVPR19 & $\mathcal{I}$ & ResNet50     & 63.5    & 64.8 \\
ICD\cite{fan2020learning}        &CVPR20 & $\mathcal{I}$ & ResNet101    & 64.1    & 64.3 \\
SCE\cite{chang2020weakly}        &CVPR20 & $\mathcal{I}$ & ResNet101    & 66.1    & 65.9 \\
SEAM\cite{wang2020self}          &CVPR20 & $\mathcal{I}$ & ResNet38     & 64.5    & 65.7 \\
BES\cite{chen2020weakly}         &ECCV20 & $\mathcal{I}$ & ResNet101    & 65.7    & 66.6 \\
MCIS\cite{sun2020mining}         &ECCV20 & $\mathcal{I}$ & ResNet101    & 66.2    & 66.9 \\
CONTA\cite{dong_2020_conta}      &NIPS20 & $\mathcal{I}$ & ResNet101    & 66.1    & 66.7 \\
LIID\cite{liu2020leveraging}     &TPAMI20 & $\mathcal{I}$ & ResNet101   & 66.5    & 67.5 \\
A2GNN\cite{zhang2021affinity}    &TPAMI21 & $\mathcal{I}$ & ResNet101   & 66.8    & 67.4 \\
AdvCAM\cite{lee2021anti}         &CVPR21 & $\mathcal{I}$ & ResNet101    & 68.1    & 68.0 \\
CDA\cite{su2021context}          &ICCV21 & $\mathcal{I}$ & ResNet38     & 66.1    & 66.8 \\
ECS\cite{sun2021ecs}             &ICCV21 & $\mathcal{I}$ & ResNet38     & 66.6    & 67.6 \\
CSE\cite{Kweon2021unlock}        &ICCV21 & $\mathcal{I}$ & ResNet38     & 68.4    & 68.2 \\
CPN\cite{zhang2021complementary} &ICCV21 & $\mathcal{I}$ & ResNet38     & 67.8    & 68.5 \\
\midrule[0.5pt]
SIPE (Ours)                      & & $\mathcal{I}$ & ResNet38    & 68.2 & 69.5 \tablefootnote {\href{http://host.robots.ox.ac.uk:8080/anonymous/NGICBM.html}{http://host.robots.ox.ac.uk:8080/anonymous/NGICBM.html}} \\
SIPE (Ours)                      & & $\mathcal{I}$ & ResNet101    & \textbf{68.8} &  \textbf{69.7} \tablefootnote
{\href{http://host.robots.ox.ac.uk:8080/anonymous/UU6VNX.html}{http://host.robots.ox.ac.uk:8080/anonymous/UU6VNX.html}} \\

\bottomrule[1pt]
\end{tabular}
\label{tab:tab2}
\end{table}}

\begin{figure*}[t]
    \centering
    \includegraphics[width=\textwidth]{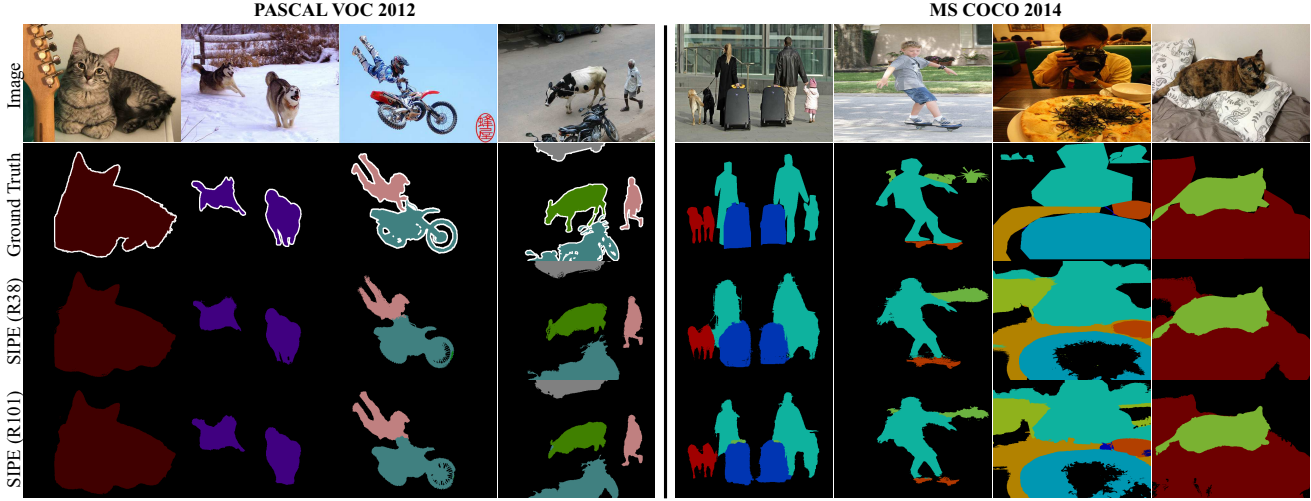}
    \caption{Qualitative segmentation results on the validation set of PASCAL VOC 2012 and MS COCO 2014.}
    \label{fig:fig6}
\end{figure*}

\renewcommand\arraystretch{1.0}
\setlength{\tabcolsep}{1.4mm}{
\begin{table}[!t]
\centering
\caption{Comparison with other state-of-the-arts on MS COCO 2014 \textit{val} set. The best results are shown in bold.}
\begin{tabular}{lcccc}
\toprule[1pt]
Model                   & Pub.      & Sup.          & Backbone   & Val         \\
\midrule[0.5pt]
DSRG\cite{huang2018weakly}  &CVPR18    & $\mathcal{I}$ + $\mathcal{S}$   & VGG16   & 26.0  \\
G-WSSS\cite{li2021group}  &AAAI21   & $\mathcal{I}$ + $\mathcal{S}$   & ResNet101   & 28.4 \\
EPS\cite{lee2021railroad}   &CVPR21    & $\mathcal{I}$ + $\mathcal{S}$   & ResNet101   & 35.7  \\
AuxSegNet\cite{xu2021leveraging} &ICCV21 & $\mathcal{I}$ + $\mathcal{S}$  & ResNet38   & 33.9  \\
\midrule[0.5pt]
SEC\cite{kolesnikov2016seed} &ECCV16   & $\mathcal{I}$   & VGG16        & 22.4          \\
IRN\cite{ahn2019weakly}  &CVPR19  & $\mathcal{I}$   & ResNet50     & 32.6          \\
IAL\cite{wang2020weakly}  &IJCV20  & $\mathcal{I}$   & VGG16    & 27.7\\
SEAM\cite{wang2020self}  &CVPR20 & $\mathcal{I}$   & ResNet38     & 31.9          \\
CONTA\cite{dong_2020_conta} &NIPS20 & $\mathcal{I}$   & ResNet101     & 33.4       \\
CSE\cite{Kweon2021unlock} &ICCV21 & $\mathcal{I}$   & ResNet38     & 36.4\\
\midrule[0.5pt]
SIPE (Ours)               &      & $\mathcal{I}$   & ResNet38    & \textbf{43.6} \\
SIPE (Ours)               &      & $\mathcal{I}$   & ResNet101    & 40.6 \\
\bottomrule[1pt]
\end{tabular}
\label{tab:tab3}
\end{table}}

\paragraph{Improvements on segmentation results.}
To further evaluate the performance of our methods, we train fully supervised models using generated pseudo labels and compare the segmentation results with the state-of-the-arts.
Following the common practice \cite{lee2021anti}, the pseudo labels are refined by IRN \cite{ahn2019weakly} and used to train DeepLabV2 \cite{chen2017deeplab}.
\cref{tab:tab2} presents the comparison with state-of-the-art methods on PASCAL VOC 2012 val and test sets.
Using only image-level labels, our SIPE outperforms previous methods with 68.8\% mIoU on val set and 69.7\% mIoU on test set.
In addition, our method performs favorably against NSROM \cite{yao2021non} and EPS \cite{lee2021railroad}, which introduce saliency maps as auxiliary labels for this task.
For a fair comparison, we also train the model with ResNet38 following the default setting as \cite{wang2020self}.
Our SIPE achieves 69.5\% mIoU on test set, exceeding the existing methods that use ResNet38 backbone.
The qualitative segmentation results on val set are shown in \cref{fig:fig6}.
Based on our SIPE, DeepLabV2 shows more robustness to various challenging scenarios, such as different object scales, multiple objects and multiple categories.

In \cref{tab:tab3}, we also evaluate our method in MS COCO 2014 dataset \cite{lin2014coco}.
The same training script with the experiment on PASCAL VOC 2012 is employed, but no refinement with IRN due to the large computation cost.
Our method achieves 43.6\% mIoU with ResNet38 backbone on the validation set, which is 7.2\% higher than previous SOTA CSE \cite{Kweon2021unlock}.
Using ResNet101 backbone, we also outperforms EPS \cite{lee2021railroad} by 4.9\%.
These outstanding performances over existing state-of-the-arts on both datasets confirm the effectiveness of our SIPE, which well explores image-specific prototypes via self-supervised learning paradigm.

\subsection{Ablation Studies}

\setlength{\tabcolsep}{2.9mm}{
\begin{table}[tb]
\caption{Effect of the main contributions. CAM: orginal CAM, IPE: image-specific prototype exploration, GSC: general-specific consistency.}
\centering
\begin{tabular}{cccc}
\toprule[1pt]
CAM              & IPE       & GSC         & mIoU (\%) \\
\midrule[0.5pt]
\checkmark       &             &             & 50.1      \\
\checkmark       & \checkmark  &             & 53.2      \\
\checkmark       & \checkmark  & \checkmark  & 58.6      \\
\bottomrule[1pt]
\end{tabular}
\label{tab:tab4}
\end{table}}

\begin{figure}[t]
\centering
\includegraphics[width=0.47\textwidth]{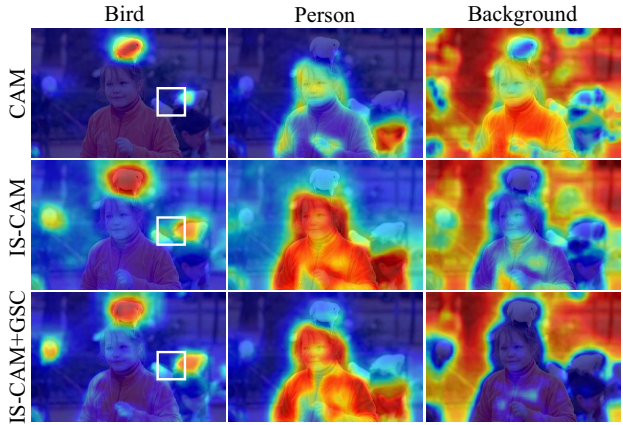}
\caption{Visualization of CAM and IS-CAM with different settings. Our IS-CAM shows more complete regions of person. Moreover, the proposed GSC further refines the background activations. The boundary is well captured as the white box.}
\label{fig:fig7}
\end{figure}

\textbf{Effect of the main contributions.}
We conduct an ablation study to verify the effect of the proposed two key contributions, i.e., Image-specific Prototype Exploration (IPE) and General-Specific Consistency (GSC).
As shown in \cref{tab:tab4}, with image-specific prototypes, our IS-CAM can improve the original CAM by 2.1\% on mIoU score.
To enhance the feature representation, GSC is introduced for self-supervised training.
The proposed GSC improves the quality of IS-CAM (5.4\%) by a clear margin.
By combining these two methods, our full method performs significantly better than the original CAM.
\cref{fig:fig7} visualizes CAM and IS-CAM with different settings.
From the first two rows, we can observe that image-specific prototypes can activate more useful regions.
Additionally, our IS-CAM shows that the proposed method produces more clear background activations than that from CAM.
When training our model with GSC, the quality of localization maps is obviously improved, especially for background.

\begin{figure}[t]
    \centering
    \includegraphics[width=0.47\textwidth]{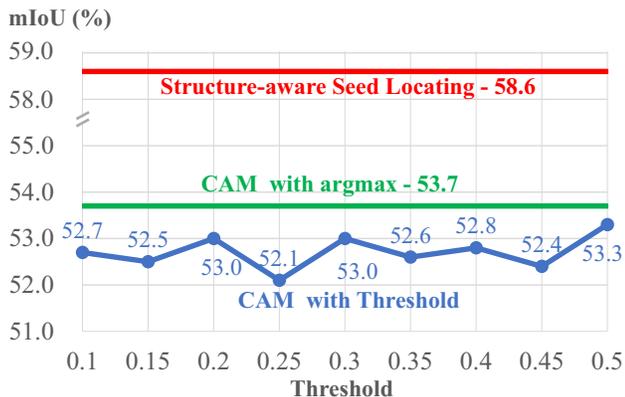}
    \caption{Ablation study of structure-aware seed locating. The proposed structural-aware seed locating method is consistently better than CAM with fixed threshold, as well as CAM with adaptive threshold.}
    \label{fig:fig8}
\end{figure}

\textbf{Effect of structure-aware seed locating.}
To verify the effectiveness of the proposed structure-aware seed locating, we compare the mIoU with other seed generation approaches including applying threshold or argmax operations on CAM.
As shown in \cref{fig:fig8}, applying different thresholds on CAM exhibits fluctuations and can only achieve maximal mIoU of 53.3\%.
In addition, simply applying argmax results in slightly performance gains (+0.4\%).
These methods are still hard to find compete regions because both of them only depend on the single pixel's probability.
In contrast, the proposed structure-aware seed locating largely outperforms the above methods owing to the proposed structure information.

\textbf{Effect of prototype modeling.}
We conduct ablation study of prototype modeling on localization maps.
The IoUs of four options concerning feature and Background Prototype Modeling (BPM) are presented in \cref{tab:tab5}:
(1) Semantic feature without BPM,
(2) Semantic feature with BPM,
(3) Hierarchical feature without BPM,
and (4) Hierarchical feature with BPM.
We report threshold-based and estimated-based pseudo labels for each type of option.
Our SIPE adopts option (4) where hierarchical feature is learned for image-specific background modeling.
It shows that our method achieves the highest performance in both values among all options.
From (1) to (2), the mIoU is basically unchanged because the background often does not have specific semantics.
From (1) to (3), the mIoU drops over 2\% since the low-level information brings confusion to foreground localization.
It is worth noting that only when hierarchical feature and background modeling are employed at the same time, the mIoU of the estimated-based pseudo labels exceed that bases on threshold, which strongly proves the effectiveness of estimating background cues.

\setlength{\tabcolsep}{0.8mm}{
\begin{table}[t]
\caption{Ablation study of combinations of feature selection and Background Prototype Modeling (BPM) via GSC. Each item reports mIoUs for two pseudo labels where the former are generated by searching the best background threshold \cite{wang2020self} and the latter are generated by estimated background map. The best results are shown in bold.}
\centering
\begin{tabular}{ccccc}
\toprule[1pt]
\multirow{2}{*}{Feature} & \multicolumn{2}{c}{w/o BPM} & \multicolumn{2}{c}{w/ BPM} \\
 \cmidrule{2-5}
& Threshold & Estimated & Threshold & Estimated\\
\midrule[0.5pt]
Semantic & 56.0 &  54.6  & 55.9  & 54.3  \\
Hierarchical & 53.9 & 51.2  & \textbf{57.6} & \textbf{58.6}\\
\bottomrule[1pt]
\end{tabular}
\label{tab:tab5}
\end{table}}

\section{Conclusion}
In this paper, we propose Self-supervised Image-specific Prototype Exploration (SIPE) for weakly supervised semantic segmentation.
In our framework, an Image-specific Prototype Exploration (IPE) is proposed to achieve more favorable localization maps.
It is achieved by structure-aware seed locating and background-aware prototype modeling.
In addition, a General-Specific Consistency (GSC) loss is developed to efficiently regularize the general CAM and Image-Specific CAM (IS-CAM), empowering the feature representation.
Extensive experiments show that our SIPE sets new state-of-the-art performance on two well-known benchmarks using image-level labels.

\paragraph{Acknowledgments.} This project is supported by the Key-Area Research and Development Program of Guangdong Province (2019B010155003), and the National Natural Science Foundation of China (62072482). We would like to thank Pengze Zhang, Huajun Zhou, and Xingxing Weng for insight discussion.

\newpage
{\small
\bibliographystyle{ieee_fullname}
\bibliography{cvpr}

\begin{thebibliography}{10}\itemsep=-1pt

\bibitem{ahn2019weakly}
Jiwoon Ahn, Sunghyun Cho, and Suha Kwak.
\newblock Weakly supervised learning of instance segmentation with inter-pixel
  relations.
\newblock In {\em Proceedings of the IEEE Conference on Computer Vision and
  Pattern Recognition}, pages 2209--2218, 2019.

\bibitem{ahn2018learning}
Jiwoon Ahn and Suha Kwak.
\newblock Learning pixel-level semantic affinity with image-level supervision
  for weakly supervised semantic segmentation.
\newblock In {\em Proceedings of the IEEE Conference on Computer Vision and
  Pattern Recognition}, pages 4981--4990, 2018.

\bibitem{bearman2016s}
Amy Bearman, Olga Russakovsky, Vittorio Ferrari, and Li Fei-Fei.
\newblock What’s the point: Semantic segmentation with point supervision.
\newblock In {\em European Conference on Computer Vision}, pages 549--565,
  2016.

\bibitem{chang2020weakly}
Yu-Ting Chang, Qiaosong Wang, Wei-Chih Hung, Robinson Piramuthu, Yi-Hsuan Tsai,
  and Ming-Hsuan Yang.
\newblock Weakly-supervised semantic segmentation via sub-category exploration.
\newblock In {\em Proceedings of the IEEE Conference on Computer Vision and
  Pattern Recognition}, pages 8991--9000, 2020.

\bibitem{Chen2021Seminar}
Hongjun Chen, Jinbao Wang, Hong~Cai Chen, Xiantong Zhen, Feng Zheng, Rongrong
  Ji, and Ling Shao.
\newblock Seminar learning for click-level weakly supervised semantic
  segmentation.
\newblock In {\em Proceedings of the IEEE International Conference on Computer
  Vision}, pages 6920--6929, October 2021.

\bibitem{chen2020weakly}
Liyi Chen, Weiwei Wu, Chenchen Fu, Xiao Han, and Yuntao Zhang.
\newblock Weakly supervised semantic segmentation with boundary exploration.
\newblock In {\em European Conference on Computer Vision}, pages 347--362,
  2020.

\bibitem{chen2017deeplab}
Liang-Chieh Chen, George Papandreou, Iasonas Kokkinos, Kevin Murphy, and Alan~L
  Yuille.
\newblock Deeplab: Semantic image segmentation with deep convolutional nets,
  atrous convolution, and fully connected crfs.
\newblock {\em IEEE Transactions on Pattern Analysis and Machine Intelligence},
  40(4):834--848, 2017.

\bibitem{choe2020attention}
Junsuk Choe, Seungho Lee, and Hyunjung Shim.
\newblock Attention-based dropout layer for weakly supervised single object
  localization and semantic segmentation.
\newblock {\em IEEE Transactions on Pattern Analysis and Machine Intelligence},
  2020.

\bibitem{deng2009imagenet}
Jia Deng, Wei Dong, Richard Socher, Li-Jia Li, Kai Li, and Li Fei-Fei.
\newblock Imagenet: A large-scale hierarchical image database.
\newblock In {\em Proceedings of the IEEE Conference on Computer Vision and
  Pattern Recognition}, pages 248--255, 2009.

\bibitem{dong_2020_conta}
Zhang Dong, Zhang Hanwang, Tang Jinhui, Hua Xiansheng, and Sun Qianru.
\newblock Causal intervention for weakly supervised semantic segmentation.
\newblock In {\em Advances in Neural Information Processing Systems}, 2020.

\bibitem{everingham2015pascal}
Mark Everingham, SM~Ali Eslami, Luc Van~Gool, Christopher~KI Williams, John
  Winn, and Andrew Zisserman.
\newblock The pascal visual object classes challenge: A retrospective.
\newblock {\em International Journal of Computer Vision}, 111(1):98--136, 2015.

\bibitem{fan2020learning}
Junsong Fan, Zhaoxiang Zhang, Chunfeng Song, and Tieniu Tan.
\newblock Learning integral objects with intra-class discriminator for
  weakly-supervised semantic segmentation.
\newblock In {\em Proceedings of the IEEE Conference on Computer Vision and
  Pattern Recognition}, pages 4283--4292, 2020.

\bibitem{fan2020cian}
Junsong Fan, Zhaoxiang Zhang, Tieniu Tan, Chunfeng Song, and Jun Xiao.
\newblock Cian: Cross-image affinity net for weakly supervised semantic
  segmentation.
\newblock In {\em Proceedings of the AAAI Conference on Artificial
  Intelligence}, volume~34, pages 10762--10769, 2020.

\bibitem{feng2020deep}
Di Feng, Christian Haase-Schuetz, Lars Rosenbaum, Heinz Hertlein, Claudius
  Glaeser, Fabian Timm, Werner Wiesbeck, and Klaus Dietmayer.
\newblock Deep multi-modal object detection and semantic segmentation for
  autonomous driving: Datasets, methods, and challenges.
\newblock {\em IEEE Transactions on Intelligent Transportation Systems}, 2020.

\bibitem{hariharan2011semantic}
Bharath Hariharan, Pablo Arbel{\'a}ez, Lubomir Bourdev, Subhransu Maji, and
  Jitendra Malik.
\newblock Semantic contours from inverse detectors.
\newblock In {\em Proceedings of the IEEE International Conference on Computer
  Vision}, pages 991--998, 2011.

\bibitem{he2016deep}
Kaiming He, Xiangyu Zhang, Shaoqing Ren, and Jian Sun.
\newblock Deep residual learning for image recognition.
\newblock In {\em Proceedings of the IEEE Conference on Computer Vision and
  Pattern Recognition}, pages 770--778, 2016.

\bibitem{hossain2019segmentation}
Mohammad~D Hossain and Dongmei Chen.
\newblock Segmentation for object-based image analysis (obia): A review of
  algorithms and challenges from remote sensing perspective.
\newblock {\em ISPRS Journal of Photogrammetry and Remote Sensing},
  150:115--134, 2019.

\bibitem{hou2018self}
Qibin Hou, Peng-Tao Jiang, Yunchao Wei, and Ming-Ming Cheng.
\newblock Self-erasing network for integral object attention.
\newblock In {\em Advances in Neural Information Processing Systems}, pages
  547--557, 2018.

\bibitem{huang2018weakly}
Zilong Huang, Xinggang Wang, Jiasi Wang, Wenyu Liu, and Jingdong Wang.
\newblock Weakly-supervised semantic segmentation network with deep seeded
  region growing.
\newblock In {\em Proceedings of the IEEE Conference on Computer Vision and
  Pattern Recognition}, pages 7014--7023, 2018.

\bibitem{jiang2019integral}
Peng-Tao Jiang, Qibin Hou, Yang Cao, Ming-Ming Cheng, Yunchao Wei, and Hong-Kai
  Xiong.
\newblock Integral object mining via online attention accumulation.
\newblock In {\em Proceedings of the IEEE International Conference on Computer
  Vision}, pages 2070--2079, 2019.

\bibitem{kim2021discriminative}
Beomyoung Kim, Sangeun Han, and Junmo Kim.
\newblock Discriminative region suppression for weakly-supervised semantic
  segmentation.
\newblock In {\em Proceedings of the AAAI Conference on Artificial
  Intelligence}, volume~35, pages 1754--1761, 2021.

\bibitem{kolesnikov2016seed}
Alexander Kolesnikov and Christoph~H. Lampert.
\newblock Seed, expand and constrain: Three principles for weakly-supervised
  image segmentation.
\newblock In {\em European Conference on Computer Vision}, 2016.

\bibitem{Kweon2021unlock}
Hyeokjun Kweon, Sung-Hoon Yoon, Hyeonseong Kim, Daehee Park, and Kuk-Jin Yoon.
\newblock Unlocking the potential of ordinary classifier: Class-specific
  adversarial erasing framework for weakly supervised semantic segmentation.
\newblock In {\em Proceedings of the IEEE International Conference on Computer
  Vision}, pages 6994--7003, October 2021.

\bibitem{lee2019ficklenet}
Jungbeom Lee, Eunji Kim, Sungmin Lee, Jangho Lee, and Sungroh Yoon.
\newblock Ficklenet: Weakly and semi-supervised semantic image segmentation
  using stochastic inference.
\newblock In {\em Proceedings of the IEEE Conference on Computer Vision and
  Pattern Recognition}, pages 5267--5276, 2019.

\bibitem{lee2021anti}
Jungbeom Lee, Eunji Kim, and Sungroh Yoon.
\newblock Anti-adversarially manipulated attributions for weakly and
  semi-supervised semantic segmentation.
\newblock In {\em Proceedings of the IEEE Conference on Computer Vision and
  Pattern Recognition}, pages 4071--4080, June 2021.

\bibitem{lee2021bbam}
Jungbeom Lee, Jihun Yi, Chaehun Shin, and Sungroh Yoon.
\newblock Bbam: Bounding box attribution map for weakly supervised semantic and
  instance segmentation.
\newblock In {\em Proceedings of the IEEE Conference on Computer Vision and
  Pattern Recognition}, pages 2643--2652, 2021.

\bibitem{lee2021railroad}
Seungho Lee, Minhyun Lee, Jongwuk Lee, and Hyunjung Shim.
\newblock Railroad is not a train: Saliency as pseudo-pixel supervision for
  weakly supervised semantic segmentation.
\newblock In {\em Proceedings of the IEEE Conference on Computer Vision and
  Pattern Recognition}, pages 5495--5505, June 2021.

\bibitem{li2021group}
Xueyi Li, Tianfei Zhou, Jianwu Li, Yi Zhou, and Zhaoxiang Zhang.
\newblock Group-wise semantic mining for weakly supervised semantic
  segmentation.
\newblock In {\em Proceedings of the AAAI Conference on Artificial
  Intelligence}, pages 1984--1992, May 2021.

\bibitem{lin2016scribblesup}
Di Lin, Jifeng Dai, Jiaya Jia, Kaiming He, and Jian Sun.
\newblock Scribblesup: Scribble-supervised convolutional networks for semantic
  segmentation.
\newblock In {\em Proceedings of the IEEE Conference on Computer Vision and
  Pattern Recognition}, pages 3159--3167, 2016.

\bibitem{lin2014coco}
Tsung-Yi Lin, Michael Maire, Serge Belongie, James Hays, Pietro Perona, Deva
  Ramanan, Piotr Doll{\'a}r, and C~Lawrence Zitnick.
\newblock Microsoft coco: Common objects in context.
\newblock In {\em European conference on computer vision}, pages 740--755,
  2014.

\bibitem{liu2020weakly}
Weide Liu, Chi Zhang, Guosheng Lin, Tzu-Yi Hung, and Chunyan Miao.
\newblock Weakly supervised segmentation with maximum bipartite graph matching.
\newblock In {\em Proceedings of the 28th ACM International Conference on
  Multimedia}, pages 2085--2094, 2020.

\bibitem{liu2020leveraging}
Yun Liu, Yu-Huan Wu, Pei-Song Wen, Yu-Jun Shi, Yu Qiu, and Ming-Ming Cheng.
\newblock Leveraging instance-, image-and dataset-level information for weakly
  supervised instance segmentation.
\newblock {\em IEEE Transactions on Pattern Analysis and Machine Intelligence},
  2020.

\bibitem{liu2020part}
Yongfei Liu, Xiangyi Zhang, Songyang Zhang, and Xuming He.
\newblock Part-aware prototype network for few-shot semantic segmentation.
\newblock In {\em European Conference on Computer Vision}, pages 142--158.
  Springer, 2020.

\bibitem{Pan2021Scribble}
Zhiyi Pan, Peng Jiang, Yunhai Wang, Changhe Tu, and Anthony~G. Cohn.
\newblock Scribble-supervised semantic segmentation by uncertainty reduction on
  neural representation and self-supervision on neural eigenspace.
\newblock In {\em Proceedings of the IEEE International Conference on Computer
  Vision}, pages 7416--7425, October 2021.

\bibitem{su2021context}
Yukun Su, Ruizhou Sun, Guosheng Lin, and Qingyao Wu.
\newblock Context decoupling augmentation for weakly supervised semantic
  segmentation.
\newblock 2021.

\bibitem{sun2020mining}
Guolei Sun, Wenguan Wang, Jifeng Dai, and Luc Van~Gool.
\newblock Mining cross-image semantics for weakly supervised semantic
  segmentation.
\newblock In {\em European Conference on Computer Vision}, pages 347--365,
  2020.

\bibitem{sun2021ecs}
Kunyang Sun, Haoqing Shi, Zhengming Zhang, and Yongming Huang.
\newblock Ecs-net: Improving weakly supervised semantic segmentation by using
  connections between class activation maps.
\newblock In {\em Proceedings of the IEEE International Conference on Computer
  Vision}, pages 7283--7292, October 2021.

\bibitem{tajbakhsh2020embracing}
Nima Tajbakhsh, Laura Jeyaseelan, Qian Li, Jeffrey~N Chiang, Zhihao Wu, and
  Xiaowei Ding.
\newblock Embracing imperfect datasets: A review of deep learning solutions for
  medical image segmentation.
\newblock {\em Medical Image Analysis}, 63:101693, 2020.

\bibitem{van2008visualizing}
Laurens Van~der Maaten and Geoffrey Hinton.
\newblock Visualizing data using t-sne.
\newblock {\em Journal of machine learning research}, 9(11), 2008.

\bibitem{wang2019panet}
Kaixin Wang, Jun~Hao Liew, Yingtian Zou, Daquan Zhou, and Jiashi Feng.
\newblock Panet: Few-shot image semantic segmentation with prototype alignment.
\newblock In {\em Proceedings of the IEEE International Conference on Computer
  Vision}, October 2019.

\bibitem{wang2020weakly}
Xiang Wang, Sifei Liu, Huimin Ma, and Ming-Hsuan Yang.
\newblock Weakly-supervised semantic segmentation by iterative affinity
  learning.
\newblock {\em International Journal of Computer Vision}, 128(6):1736--1749,
  2020.

\bibitem{wang2020self}
Yude Wang, Jie Zhang, Meina Kan, Shiguang Shan, and Xilin Chen.
\newblock Self-supervised equivariant attention mechanism for weakly supervised
  semantic segmentation.
\newblock In {\em Proceedings of the IEEE Conference on Computer Vision and
  Pattern Recognition}, pages 12275--12284, 2020.

\bibitem{wei2017object}
Yunchao Wei, Jiashi Feng, Xiaodan Liang, Ming-Ming Cheng, Yao Zhao, and
  Shuicheng Yan.
\newblock Object region mining with adversarial erasing: A simple
  classification to semantic segmentation approach.
\newblock In {\em Proceedings of the IEEE Conference on Computer Vision and
  Pattern Recognition}, pages 1568--1576, 2017.

\bibitem{wei2018revisiting}
Yunchao Wei, Huaxin Xiao, Honghui Shi, Zequn Jie, Jiashi Feng, and Thomas~S
  Huang.
\newblock Revisiting dilated convolution: A simple approach for weakly-and
  semi-supervised semantic segmentation.
\newblock In {\em Proceedings of the IEEE Conference on Computer Vision and
  Pattern Recognition}, pages 7268--7277, 2018.

\bibitem{wu2021embed}
Tong Wu, Junshi Huang, Guangyu Gao, Xiaoming Wei, Xiaolin Wei, Xuan Luo, and
  Chi~Harold Liu.
\newblock Embedded discriminative attention mechanism for weakly supervised
  semantic segmentation.
\newblock In {\em Proceedings of the IEEE Conference on Computer Vision and
  Pattern Recognition}, pages 16765--16774, June 2021.

\bibitem{Xu2021Scribble}
Jingshan Xu, Chuanwei Zhou, Zhen Cui, Chunyan Xu, Yuge Huang, Pengcheng Shen,
  Shaoxin Li, and Jian Yang.
\newblock Scribble-supervised semantic segmentation inference.
\newblock In {\em Proceedings of the IEEE International Conference on Computer
  Vision}, pages 15354--15363, October 2021.

\bibitem{xu2021leveraging}
Lian Xu, Wanli Ouyang, Mohammed Bennamoun, Farid Boussaid, Ferdous Sohel, and
  Dan Xu.
\newblock Leveraging auxiliary tasks with affinity learning for weakly
  supervised semantic segmentation.
\newblock In {\em Proceedings of the IEEE International Conference on Computer
  Vision}, 2021.

\bibitem{yao2021non}
Yazhou Yao, Tao Chen, Guo-Sen Xie, Chuanyi Zhang, Fumin Shen, Qi Wu, Zhenmin
  Tang, and Jian Zhang.
\newblock Non-salient region object mining for weakly supervised semantic
  segmentation.
\newblock In {\em Proceedings of the IEEE Conference on Computer Vision and
  Pattern Recognition}, pages 2623--2632, June 2021.

\bibitem{zhang2021affinity}
Bingfeng Zhang, Jimin Xiao, Jianbo Jiao, Yunchao Wei, and Yao Zhao.
\newblock Affinity attention graph neural network for weakly supervised
  semantic segmentation.
\newblock {\em IEEE Transactions on Pattern Analysis and Machine Intelligence},
  2021.

\bibitem{zhang2020reliability}
Bingfeng Zhang, Jimin Xiao, Yunchao Wei, Mingjie Sun, and Kaizhu Huang.
\newblock Reliability does matter: An end-to-end weakly supervised semantic
  segmentation approach.
\newblock In {\em Proceedings of the AAAI Conference on Artificial
  Intelligence}, volume~34, pages 12765--12772, 2020.

\bibitem{zhang2021complementary}
Fei Zhang, Chaochen Gu, Chenyue Zhang, and Yuchao Dai.
\newblock Complementary patch for weakly supervised semantic segmentation.
\newblock In {\em Proceedings of the IEEE International Conference on Computer
  Vision}, 2021.

\bibitem{zhang2020sg}
Xiaolin Zhang, Yunchao Wei, Yi Yang, and Thomas~S Huang.
\newblock Sg-one: Similarity guidance network for one-shot semantic
  segmentation.
\newblock {\em IEEE transactions on cybernetics}, 50(9):3855--3865, 2020.

\bibitem{Zhou_2016_CVPR}
Bolei Zhou, Aditya Khosla, Agata Lapedriza, Aude Oliva, and Antonio Torralba.
\newblock Learning deep features for discriminative localization.
\newblock In {\em Proceedings of the IEEE Conference on Computer Vision and
  Pattern Recognition}, pages 2921--2929, June 2016.

\end{thebibliography}
}

\end{document}